%% file: emnlp2021.tex
\pdfoutput=1
\documentclass[11pt]{article}

\usepackage{emnlp2021}

\usepackage{times}
\usepackage{latexsym}

\usepackage[T1]{fontenc}
\usepackage[utf8]{inputenc}

\usepackage{microtype}

\usepackage[]{algorithm2e}
\usepackage{algorithmic,float}
\usepackage[utf8]{inputenc} % allow utf-8 input
\usepackage[T1]{fontenc}    % use 8-bit T1 fonts
\usepackage{booktabs}       % professional-quality tables
\usepackage{amsfonts}       % blackboard math symbols
\usepackage{nicefrac}       % compact symbols for 1/2, etc.
\usepackage{microtype}      % microtypography
\usepackage{multirow}
\usepackage{caption}
\usepackage{amsmath,amssymb}
\usepackage{dsfont}			% make hollow 1 work
\usepackage{cases}
\usepackage{tikz}
\usepackage{pgfplots, pgfplotstable}
\usetikzlibrary{shapes,arrows,calc}
\usepackage{caption}
\usepackage{textcomp}
\usepackage{enumitem}
\usepackage{import}
\usepackage{CJKutf8}
\usepackage[toc,page]{appendix}
\usepackage[textsize=large]{todonotes}
\usepackage[normalem]{ulem}
\usepackage[scaled=.8]{beramono}
\usepackage{fancyvrb}
\useunder{\uline}{\ul}{}

\setlist[itemize]{leftmargin=15pt}
\newcommand\blfootnote[1]{%
  \begingroup
  \renewcommand\thefootnote{}\footnote{#1}%
  \addtocounter{footnote}{-1}%
  \endgroup
}

\title{The JHU-Microsoft Submission for\\ WMT21 Quality Estimation Shared Task}

\author{Shuoyang Ding$\dagger^*$\quad Marcin Junczys-Dowmunt$\ddagger$\\ {\bf Matt Post$\dagger$$\ddagger$\quad Christian Federmann$\ddagger$\quad Philipp Koehn$\dagger$} \\
$\dagger$ Center for Language and Speech Processing, Johns Hopkins University\quad $\ddagger$ Microsoft \\
\texttt{\{dings, phi\}@jhu.edu}\quad \texttt{\{marcin.junczysdowmunt, mattpost, chrife\}@microsoft.com}
}

\begin{document}
\maketitle
\begin{abstract}
This paper presents the JHU-Microsoft joint submission for WMT 2021 quality estimation shared task.
We only participate in Task 2 (post-editing effort estimation) of the shared task, focusing on the target-side word-level quality estimation.
The techniques we experimented with include Levenshtein Transformer training and data augmentation with a combination of forward, backward, round-trip translation, and pseudo post-editing of the MT output.
We demonstrate the competitiveness of our system compared to the widely adopted OpenKiwi-XLM baseline.
Our system is also the top-ranking system on the MT MCC metric for the English-German language pair.
\end{abstract}

\import{\sectiondir}{intro.tex}
\import{\sectiondir}{method.tex}
\import{\sectiondir}{experiment.tex}
\import{\sectiondir}{conclusion.tex}

\bibliography{anthology,custom}
\bibliographystyle{acl_natbib}

% \appendix
% \section{Example Appendix}
% \label{sec:appendix}
% This is an appendix.

\end{document}

%% file: source/intro.tex
\section{Introduction\blfootnote{$^*$ Shuoyang Ding had a part-time affiliation with Microsoft at the time of this work.}}

In the machine translation (MT) literature, quality estimation (QE) refers to the task of evaluating the translation quality of a system without using a human-generated reference.
There are several different granularities as to the way those quality labels or scores are generated.
Our participation in the WMT21 quality estimation shared task focuses specifically on the \emph{word-level} quality labels (word-level subtask of Task 2), which are helpful for both human \cite{lee-etal-2021-intellicat} and automatic \cite{lee-2020-cross} post-editing of translation outputs.
The task asks the participant to predict one binary quality label (\texttt{OK}/\texttt{BAD}) for each target word and each gap between target words, respectively.\footnote{While there is another sub-task for predicting source-side quality labels, we do not participate in that task.}

Our approach closely follows our contemporary work \cite{ding2021levenshtein}, which focuses on en-de and en-zh language pairs tested in the 2020 version of the shared task.
The intuition behind our idea is that translation knowledge is very useful for predicting word-level quality labels of translations.
However, usage of machine translation models is limited in the previous work mainly due to (1) the difficulties in using both the left and right context of an MT word to be evaluated;
(2) the difficulties in making the word-level reference labels compatible with subword-level models;
and (3) the difficulties in enabling translation models to predict gap labels.
To resolve these difficulties, we resort to Levenshtein Transformer \cite[LevT,][]{DBLP:conf/nips/GuWZ19}, a model architecture designed for non-autoregressive neural machine translation (NA-NMT).
Because of its iterative inference procedure, LevT is capable of performing post-editing on existing translation output even just trained for translation.
To further improve the model performance, we also propose to initialize the encoder and decoder of the LevT model with those from a massively pre-trained multilingual NMT model \cite[M2M-100,][]{DBLP:journals/corr/abs-2010-11125}.

Starting from a LevT translation model, we then perform a two-stage finetuning process to adapt the model from translation prediction to quality label prediction, using automatically-generated pseudo-post-editing triplets and human post-editing triplets respectively.
% We experimented with a series of data augmentation techniques, including forward translation, round-trip translation, and a novel zero-shot pseudo-post-editing scheme with a multilingual translation model.
All of our final system submissions are also linear ensembles from several individual models with weights optimized on the development set using the Nelder-Mead method \cite{DBLP:journals/cj/NelderM65}.

%% file: source/method.tex
\section{Method}

Our system building pipeline is consisted of three different stages:
\begin{itemize} \itemsep -2pt
	\item \textbf{Stage 1}: Training LevT for translation
	\item \textbf{Stage 2 (Optional)}: Finetuning LevT on synthetic post-editing triplets
	\item \textbf{Stage 3}: Finetuning LevT on human post-editing triplets
\end{itemize}

\paragraph{Stage 1: Training LevT for Translation}
We largely follow the same procedure as \citet[LevT,][]{DBLP:conf/nips/GuWZ19} to train the LevT translation model, except that we initialize the embedding, the encoder, and decoder of LevT with those from the small M2M-100-small model \cite[418M parameters,][]{DBLP:journals/corr/abs-2010-11125} to take advantage of large-scale pretraining.
Because of that, we also use the same sentencepiece model and vocabulary as the M2M-100 model.

For to-English language pairs, we explored training multi-source LevT model. According to the results on devtest data, this is shown to be beneficial for the QE task for ro-en, ru-en and ne-en, but not for other language pairs.

\paragraph{Stage 2: Synthetic Finetuning}
During both finetuning stages, we update the model parameters to minimize the NLL loss of word quality labels and gap quality labels, for the deletion and insertion head, respectively.
% For language pairs with high translation quality, the translation errors are often quite scarce, thus creating a skewed label distribution over the \texttt{OK} and \texttt{BAD} labels.
% To ensure that the model captures both label categories in a balanced manner, we introduce a multiplicative label balance factor $\sigma$ to up-weight the NLL loss of the \texttt{BAD} labels.
To obtain training targets for finetuning, we need \emph{translation triplet data}, i.e., the aligned triplet of source, target, and post-edited segments.
Human post-edited data naturally provides all three fields of the triplet, but only comes in a limited quantity.
To further help the model to generalize, we conduct an extra step of finetuning on synthetic translation triplets, similar to some previous work \cite[][\textit{inter alia}]{lee-2020-two}.
We explored five different methods for data synthesis, namely:
\begin{enumerate}[leftmargin=0.5cm]
	\item \texttt{src-mt-tgt}: Take the source side of a parallel corpus (\texttt{src}), translate it with a MT model to obtain the MT output (\texttt{mt}), and use the target side of the parallel corpus as the pseudo post-edited output (\texttt{tgt}).
	\item \texttt{src-mt1-mt2}: Take a corpus in the source language (\texttt{src}) and translate it with two different MT systems that have clear system-level translation quality orderings. Then, take the worse MT output as the MT output in the triplet (\texttt{mt1}) and the better as the pseudo post-edited output in the triplet (\texttt{mt2}).
	\item \texttt{bt-rt-tgt}: Take a corpus in the target language (\texttt{tgt}) and back-translate it into the source langauge (\texttt{bt}), and then translate again to the target language (\texttt{rt}). We then use \texttt{rt} as the MT output in the triplet and \texttt{tgt} as the pseudo post-edited output in the triplet.
	% \item \texttt{bt-noisy-tgt}: Take a corpus in the target language (\texttt{tgt}) and back-translate it into the source language (\texttt{bt}). We then randomly mask some words in the target and let a LevT translation model complete the translation by referring to the \texttt{bt} and masked \texttt{tgt}, which will generate a noisy target sentence (\texttt{noisy}). We then use \texttt{noisy} as the MT output in the triplet and the original target sentence in the corpus (\texttt{tgt}) as the pseudo post-edited output in the triplet.
	\item \texttt{src-rt-ft}: Take a parallel corpus and translate its source side and use it as the pseudo post-edited output (\texttt{ft}), and round-trip translate its target side (\texttt{rt}) as the MT output in the translation triplet.
	\item \texttt{Multi-view Pseudo Post-Editing (MVPPE)}: Same as \citet{ding2021levenshtein}, we take a parallel corpus and translate the source side (\texttt{src}) with a multilingual translation system (\texttt{mt}) as the MT output in the triplet. We then generate the pseudo-post-edited output by ensembling two different \emph{views} of the same model: (1) using the multilingual translation model as a translation model, with \texttt{src} as the input; (2) using the multilingual translation model as a paraphrasing model, with \texttt{tgt} as the input. The ensemble process is the same as ensembling standard MT models, and we perform beam search on top of the ensemble. Unless otherwise specified, we use the same ensembling weights of $\lambda_t = 2.0$ and $\lambda_p = 1.0$ as \citet{ding2021levenshtein}.
\end{enumerate}

\paragraph{Stage 3: Human Post-editing Finetuning}
We follow the same procedure as stage 2, except that we finetune on the human post-edited dataset provided by the shared task organizers for this stage.

\begin{table*}[]
\centering
\scalebox{0.9}{
\begin{tabular}{@{}lll@{}}
\toprule
Language Pair    & Data Source                                  & Sentence Pairs \\ \midrule
English-German   & WMT20 en-de parallel data                    & 44.2M          \\
English-Chinese  & shared task en-zh parallel                   & 20.3M          \\
Romanian-English & shared task ro-en parallel                   & 3.09M          \\
Russian-English  & shared task ru-en parallel                   & 2.32M          \\
Estonian-English & shared task et-en parallel                   & 880K           \\
Estonian-English & shared task et-en parallel + NewsCrawl 14-17 & 3.42M          \\
Nepalese-English & shared task ne-en parallel                   & 498K           \\
Pashto-English   & WMT20 Parallel Corpus Filtering Task         & 347K           \\ \bottomrule
\end{tabular}
}
\caption{Source and statistics of parallel datasets used in our experiments. \label{tab:data}}
\end{table*}

\paragraph{Compatibility With Subwords}

As pointed out before, since LevT predicts edits on a subword-level starting from translation training, we must construct reference tags that are compatible with the subword segmentation done for both the MT and the post-edited output.
Specifically, we need to: (1) for inference, convert subword-level tags predicted by the model to word-level tags for evaluation, and (2) for both finetuning stages, build subword-level reference tags.
We follow the same heuristic subword-level tag reference construction procedure as \citet{ding2021levenshtein}, which was shown to be helpful for task performance.

\paragraph{Label Imbalance}
Like several previous work \cite{lee-2020-cross,wang-etal-2020-hw-tscs,moura-etal-2020-ist}, we also observed that the translation errors are often quite scarce, thus creating a skewed label distribution over the \texttt{OK} and \texttt{BAD} labels.
Since it is critical for the model to reliably predict both classes of labels, we introduce an extra hyperparameter $\mu$ in the loss function that allows us to upweight the words that are classified with \texttt{BAD} tags in the reference.
$$\mathcal{L} = \mathcal{L}_{OK} + \mu \mathcal{L}_{BAD}$$

\paragraph{Ensemble}
For each binary label prediction made by the model, the model will give a score $p(\texttt{OK})$, which are translated into binary labels in post-processing.
To ensemble predictions from $k$ models $p_1(\texttt{OK}), p_2(\texttt{OK}), \dots, p_k(\texttt{OK})$, we perform a linear combination of the scores for each label:
$$p_E(\texttt{OK}) = \lambda_1 p_1(\texttt{OK}) + \lambda_2 p_2(\texttt{OK}) + \dots + \lambda_k p_k(\texttt{OK})$$
To determine the optimal interpolation weights, we optimize towards target-side MCC on the development set.
Because the target-side MCC computation is not implemented in a way such that gradient information can be easily obtained, we experimented with two gradient-free optimization method: Powell method \cite{DBLP:journals/cj/Powell64} and Nelder-Mead method \cite{DBLP:journals/cj/NelderM65}, both as implemented in \texttt{SciPy} \cite{2020SciPy-NMeth}.
We found that the Nelder-Mead method finds better optimum on the development set while also leading to better performance on the devtest dataset (not involved in optimization).
Hence, we use the Nelder-Mead optimizer for all of our final submissions with ensembles.
We set the initial points of Nelder-Mead optimization to be the vertices of the standard simplex in the $k$-dimensional space, with $k$ being the number of the models.

We find that it is critical to build ensembles from models that yield diverse yet high-quality outputs.
Specifically, we notice that ensembles from multiple checkpoints of a single experimental run are not helpful.
Hence, for each language pair, we select 2-8 models with different training configurations that also have the highest performance to build our final ensemble model for submission.

%% file: source/experiment.tex
\section{Experiments}

% \subsection{Setup}

% \begin{figure*}
% \centering
% \includegraphics[scale=0.6]{figures/main.pdf}
% \caption{Target MCC results on \texttt{test20} dataset for all language pairs we submitted systems for (except for ps-en which is not included in \texttt{test20}).\label{fig:main}}
% \end{figure*}

\subsection{Data Setup}

\paragraph{LevT Training}
We used the same parallel data that was used to train the MT system in the shared task, except for the en-de, et-en, and ps-en language pairs.
For en-de language pair, we use the larger parallel data from the WMT20 news translation shared task.
For et-en language pair, we experiment with augmenting with the News Crawl Estonian monolingual data from 2014 to 2017, which was inspired by \citet{zhou-keung-2020-improving}.
For ps-en language pair, because there is no MT system provided, we take the data from the WMT20 parallel corpus filtering shared task and applied the baseline LASER filtering method.
For the multi-source LevT model, we simply concatenate the data from ro-en, ru-en, es-en (w/o monolingual augmentation) and ne-en.
The resulting data scale is summarized in Table \ref{tab:data}.

Following the setup in \citet{DBLP:conf/nips/GuWZ19}, we conduct sequence-level knowledge distillation during training for all language pairs except for ne-en and ps-en\footnote{The exception was motivated by the poor quality of the translation we obtained from the M2M-100 model.}.
For en-de, the knowledge distillation data is generated by the WMT19 winning submission for that language pair from Facebook \cite{ng-etal-2019-facebook}.
For en-zh, we train our own en-zh autoregressive model on the parallel data from the WMT17 news translation shared task.
For the other language pairs, we use the decoding output from M2M-100-mid (1.2B parameters) model to perform knowledge distillation.

\begin{table}[h]
\centering
\scalebox{0.68}{
\begin{tabular}{@{}lllr@{}}
\toprule
\textbf{Configuration}        & \textbf{Stage 2}      & \textbf{Stage 3} & \multicolumn{1}{l}{\textbf{Target MCC}} \\ \midrule
en-de OpenKiwi                & N                     & default          & 0.337                                   \\
en-de bilingual best          & src-mt1-mt2           & $\mu = 1.0$      & 0.500                                   \\
en-de ensemble                & N/A                   & N/A              & 0.504                                   \\ \midrule
en-zh OpenKiwi                & N                     & default          & 0.421                                   \\
en-zh bilingual best          & mvppe                 & $\mu = 1.0$      & 0.459                                   \\
en-zh ensemble                & N/A                   & N/A              & 0.466                                   \\ \midrule
ro-en OpenKiwi                & N                     & default          & 0.556                                   \\
ro-en bilingual best          & src-rt-ft             & $\mu = 1.0$      & 0.604                                   \\
ro-en multilingual best       & N                     & $\mu = 1.0$      & 0.612                                   \\
ro-en ensemble                & N/A                   & N/A              & 0.633                                   \\ \midrule
ru-en OpenKiwi                & N                     & default          & 0.279                                   \\
ru-en bilingual best          & src-rt-ft             & $\mu = 3.0$      & 0.316                                   \\
ru-en multilingual best       & N                     & $\mu = 3.0$      & 0.339                                   \\
ru-en ensemble                & N/A                   & N/A              & 0.349                                   \\ \midrule
et-en OpenKiwi                & N                     & default          & 0.503                                   \\
et-en bilingual best          & N                     & $\mu = 3.0$      & 0.556                                   \\
et-en bilingual best (w/ aug) & N                     & $\mu = 3.0$      & 0.548                                   \\
et-en multilingual best       & N                     & $\mu = 3.0$      & 0.533                                   \\
et-en ensemble                & N/A                   & N/A              & 0.575                                   \\ \midrule
ne-en OpenKiwi                & N                     & default          & 0.664                                   \\
ne-en bilingual best          & N                     & $\mu = 3.0$      & 0.677                                   \\
ne-en multilingual best       & N                     & $\mu = 3.0$      & 0.681                                   \\
ne-en ensemble                & N/A                   & N/A              & 0.688                                   \\ \bottomrule
\end{tabular}
}
\caption{Target MCC results on \texttt{test20} dataset for all language pairs we submitted systems for (except for ps-en which is not included in \texttt{test20}). Stage 2 stands for synthetic finetuning (where N stands for not performing this stage). Stage 3 stands for human annotation finetuning. $\mu$ stands for the label balancing factor.\label{tab:main}}
\end{table}

\begin{table}[h]
\centering
\scalebox{0.9}{
\begin{tabular}{@{}lrrr@{}}
\toprule
            & \multicolumn{1}{l}{\textbf{Target MCC}} & \multicolumn{1}{l}{\textbf{F1-OK}} & \multicolumn{1}{l}{\textbf{F1-BAD}} \\ \midrule
N           & 0.489                                   & 0.955                              & 0.533                               \\
src-mt-ref  & 0.493                                   & 0.955                              & 0.537                               \\
src-mt1-mt2 & \textbf{0.500}                          & 0.956                              & \textbf{0.544}                      \\
bt-rt-tgt   & 0.490                                   & 0.956                              & 0.534                               \\
src-rt-ft   & 0.494                                   & 0.956                              & 0.538                               \\
mvppe       & \textbf{0.500}                          & \textbf{0.960}                     & 0.540                               \\ \bottomrule
\end{tabular}
}
\caption{Analysis of different data synthesis methods on en-de language pair. All models here are initialized with M2M-100-small. \label{tab:data-synth-de}}
\end{table}

\begin{table*}[h]
\centering
\scalebox{0.9}{
\begin{tabular}{@{}lllrrr@{}}
\toprule
\textbf{Configuration}    & \textbf{Stage 2}       & \textbf{Stage 3} & \multicolumn{1}{l}{\textbf{Target MCC}} & \multicolumn{1}{l}{\textbf{F1-OK}} & \multicolumn{1}{l}{\textbf{F1-BAD}} \\ \midrule
ro-en multilingual        & N                      & $\mu = 1.0$      & \textbf{0.612}                   & 0.949                              & \textbf{0.659}                      \\
ro-en multilingual        & mvppe                  & $\mu = 1.0$      & 0.611                            & \textbf{0.951}                     & \textbf{0.659}                      \\
ro-en multilingual        & src-mt1-mt2 (Bing mt2) & $\mu = 1.0$      & 0.585                            & 0.936                              & 0.630                               \\
ro-en bilingual (Bing KD) & N                      & $\mu = 1.0$      & 0.581                            & 0.949                              & 0.632                               \\
ro-en bilingual (Bing KD) & src-mt1-mt2 (Bing mt2) & $\mu = 1.0$      & 0.568                            & 0.938                              & 0.619                               \\ \midrule
et-en bilingual           & N                      & $\mu = 3.0$      & 0.548                            & 0.914                              & 0.622                               \\
et-en bilingual           & mvppe                  & $\mu = 3.0$      & 0.544                            & \textbf{0.929}                     & 0.615                               \\
et-en bilingual           & src-mt1-mt2 (Bing mt2) & $\mu = 3.0$      & \textbf{0.563}                   & 0.919                              & \textbf{0.634}                      \\
et-en bilingual (Bing KD) & N                      & $\mu = 3.0$      & 0.557                            & 0.918                              & 0.629                               \\
et-en bilingual (Bing KD) & src-mt1-mt2 (Bing mt2) & $\mu = 3.0$      & 0.559                            & 0.916                              & 0.631                               \\ \bottomrule
\end{tabular}
}
\caption{Analysis of \texttt{src-mt1-mt2} and \texttt{mvppe} method on ro-en and et-en language pair. \label{tab:data-synth-ro-et}}
\end{table*}

\paragraph{Synthetic Finetuning}
We always conduct data synthesis based on the same parallel data that was used to train the LevT translation model.
For the only language pair (en-de) where we applied the \texttt{src-mt1-mt2} synthetic finetuning for shared task submission, we again use the WMT19 Facebook's winning system \cite{ng-etal-2019-facebook} to generate the higher-quality translation \texttt{mt2}, and the system provided by the shared task to generate the MT output in the pseudo translation triplet \texttt{mt1}.
For all other combinations of translation directions, language pairs and MVPPE decoding, we use the M2M-100-mid (1.2B parameters) model.

\paragraph{Human Annotation Finetuning}
We follow the data split for human post-edited data as determined by the task organizers and use \texttt{test20} as the devtest for our system development purposes.

\paragraph{Reference Tag Generation}
We implemented another TER computation tool\footnote{\texttt{https://github.com/marian-nmt/moses-scorers}} to generate the word-level and subword-level tags that we use as the reference for finetuning, but stick to the original reference tags in the test set for evaluation to avoid potential result mismatch.

\subsection{Model Setup}
Our LevT-QE model is implemented based on Fairseq \cite{DBLP:conf/naacl/OttEBFGNGA19}.
All of our experiments uses Adam optimizer \cite{DBLP:journals/corr/KingmaB14} with linear warmup and \texttt{inverse-sqrt} scheduler.
For stage 1, we use the same hyperparameters as \citet{DBLP:conf/nips/GuWZ19} for LevT translation training, but use a smaller learning rate of 2e-5 to avoid overfitting for all to-English language pairs.
For stage 2 and beyond, we stick to the learning rate of 2e-5 and perform early-stopping based on the loss function on the development set. For stage 3, e also experiment with label balancing factor $\mu=1.0$ and $\mu=3.0$ for each language pair and pick the one that works the best on devtest data, while for stage 2 we keep $\mu = 1.0$ because early experiments indicate that using $\mu = 3.0$ at this stage is not helpful.
% We find the QE task performance to be quite sensitive to $\mu$ but there is also no universally optimal value for all language pairs.

For pre-submission developments, we built OpenKiwi-XLM baselines \cite{kepler-etal-2019-openkiwi} following their \texttt{xlmroberta.yaml} recipe.
Keep in mind due to the fact that this baseline model is initialized with a much smaller XLM-Roberta-base model (281M parameters) compared to our M2M-100-small initialization (484M parameters), the performance comparison is not a strict one.

\subsection{Devtest Results}

Our system development results on \texttt{test20} devtest data are shown in Table \ref{tab:main}\footnote{Note that the results on en-zh also reflect a crucial bug fix on our TER computation tool that we added after the system submission deadline. Hence the results shown here are from a different system as in the official shared task results. The bug fix should not affect the results of the other language pairs.}.
In all language pairs, our systems can outperform the OpenKiwi baseline based upon the pre-trained XLM-RoBERTa-base encoder.
Among these language pairs, the benefit of LevT is most significant on the language pairs with a large amount of available parallel data.
Such behavior is expected, because the less parallel data we have, the less knowledge we can extract from the LevT training process.
Furthermore, the lack of good quality knowledge distillation data in the low-resource language pairs also expands this performance gap.
% In the extreme case where we have no parallel data at all, all of our knowledge comes from the pre-trained model and human annotation data finetuning, which will reduce to the same model as the baseline.
To our best knowledge, this is also the first attempt to train non-autoregressive translation systems under low-resource settings, and we hope future explorations in this area can enable us to build a better QE system from LevT.

In terms of comparison between multilingual and bilingual models for to-English language pairs, the results are mixed, with the multilingual model performing significantly better for ru-en language pair, but significantly worse for et-en language pair.
Finally, our Nelder-Mead ensemble further improves the result by a small but steady margin.

\subsection{Analysis}

\citet{ding2021levenshtein} already conducted comprehensive ablation studies for techniques such as the effect of LevT training step, heuristic subword-level reference tag, as well as the effect of various data synthesis methods.
In this section, we extend the existing analyses by studying if the synthetic finetuning is still useful with M2M initialization, and if it is universally helpful across different languages.
We also examine the effect of label balancing factor $\mu$ and take a detailed look at the prediction errors.

\paragraph{Synthetic Finetuning}
We redo the analysis on en-de synthetic finetuning with the smaller 2M parallel sentence samples from Europarl, as in \citet{ding2021levenshtein}, but with the updated \texttt{test20} test set and models with M2M-100-small initialization.
The results largely corroborate the trend in the other paper, showing that \texttt{src-mt1-mt2} and \texttt{mvppe} being the most helpful two data synthesis methods.
We then extend those two most helpful methods to ro-en and et-en, using the up-to-date Bing Translator production model as the stronger MT system (a.k.a. \texttt{mt2}) in the \texttt{src-mt1-mt2} synthetic data.
The result is mixed, with \texttt{mvppe} failing to improve performance for both language pairs, and \texttt{src-mt1-mt2} only being helpful for et-en language pair.
We also trained two extra ro-en and et-en LevT models using the respective Bing Translator models to generate the KD data, which are neither helpful for improving performance on their own nor working better with \texttt{src-mt1-mt2} synthetic data.

We notice that the \texttt{mvppe} synthetic data seems to significantly improve the F1 score of the \texttt{OK} label in general, for which we don't have a good explanation yet.

\paragraph{Label Balancing Factor}

\begin{table}[]
\scalebox{0.9}{
\begin{tabular}{@{}lrrr@{}}
\toprule
\textbf{Configuration} & \multicolumn{1}{l}{\textbf{Target MCC}} & \multicolumn{1}{l}{\textbf{F1-OK}} & \multicolumn{1}{l}{\textbf{F1-BAD}} \\ \midrule
ro-en $\mu = 1.0$      & \textbf{0.612}                          & \textbf{0.949}                     & \textbf{0.659}                      \\
ro-en $\mu = 3.0$      & 0.577                                   & 0.930                              & 0.619                               \\ \midrule
ru-en $\mu = 1.0$      & 0.267                                   & \textbf{0.960}                     & 0.284                               \\
ru-en $\mu = 3.0$      & \textbf{0.339}                          & 0.943                              & \textbf{0.390}                      \\ \midrule
et-en $\mu = 1.0$      & 0.478                                   & \textbf{0.933}                     & 0.511                               \\
et-en $\mu = 3.0$      & \textbf{0.512}                          & 0.925                              & \textbf{0.587}                      \\ \midrule
ne-en $\mu = 1.0$      & 0.660                                   & \textbf{0.885}                     & 0.774                               \\
ne-en $\mu = 3.0$      & \textbf{0.681}                          & 0.855                              & \textbf{0.788}                      \\ \bottomrule
\end{tabular}
}
\caption{Analysis of different label balancing factors initialized on to-English language pairs. All results are based on the multilingual model and not performing synthetic finetuning step. \label{tab:lb-factor}}
\end{table}

\definecolor{applegreen}{rgb}{0.553, 0.714, 0.0}
\definecolor{msgreen}{rgb}{0.490, 0.718, 0.0}
\definecolor{bananamania}{rgb}{0.98, 0.91, 0.71}
\definecolor{pastelgreen}{rgb}{0.467, 0.867, 0.467}
\def\cca#1{
    \pgfmathsetmacro\calc{(#1*100}
    \edef\clrmacro{\noexpand\cellcolor{msgreen!\calc}}
    \clrmacro{#1}
    % \ifdim \calc pt>50pt\color{white}\fi{#1}
}

\begin{table*}[]
\centering
\scalebox{0.6}{
\begin{tabular}{@{}lrrrrrrrrrrrrrrr@{}}
\toprule
\textbf{Lang.} & \multicolumn{1}{l}{\textbf{Tgt. MCC}} & \multicolumn{1}{l}{\textbf{MT MCC}} & \multicolumn{3}{l}{\textbf{MT BAD (P/R/F1)}} & \multicolumn{3}{l}{\textbf{MT OK (P/R/F1)}} & \multicolumn{1}{l}{\textbf{GAP MCC}} & \multicolumn{3}{l}{\textbf{GAP BAD (P/R/F1)}} & \multicolumn{3}{l}{\textbf{GAP OK (P/R/F1)}} \\ \midrule
en-de         &  \cca{0.504} &  \cca{0.503} &  \cca{0.476} &  \cca{0.731} &  \cca{0.576} &  \cca{0.950} &  \cca{0.863} &  \cca{0.904} &  \cca{0.280} &  \cca{0.366} &  \cca{0.238} &  \cca{0.288} &  \cca{0.980} &  \cca{0.989} &  \cca{0.984}        \\
en-zh         &  \cca{0.466} &  \cca{0.381} &  \cca{0.467} &  \cca{0.787} &  \cca{0.586} &  \cca{0.879} &  \cca{0.633} &  \cca{0.736} &  \cca{0.146} &  \cca{0.276} &  \cca{0.099} &  \cca{0.145} &  \cca{0.965} &  \cca{0.990} &  \cca{0.977}        \\
ro-en         &  \cca{0.612} &  \cca{0.645} &  \cca{0.729} &  \cca{0.709} &  \cca{0.719} &  \cca{0.922} &  \cca{0.929} &  \cca{0.926} &  \cca{0.164} &  \cca{0.411} &  \cca{0.073} &  \cca{0.125} &  \cca{0.973} &  \cca{0.997} &  \cca{0.985}        \\
ru-en         &  \cca{0.349} &  \cca{0.329} &  \cca{0.296} &  \cca{0.675} &  \cca{0.411} &  \cca{0.945} &  \cca{0.775} &  \cca{0.852} &  \cca{0.167} &  \cca{0.265} &  \cca{0.123} &  \cca{0.168} &  \cca{0.978} &  \cca{0.991} &  \cca{0.985}        \\
et-en         &  \cca{0.575} &  \cca{0.553} &  \cca{0.676} &  \cca{0.681} &  \cca{0.679} &  \cca{0.875} &  \cca{0.873} &  \cca{0.874} &  \cca{0.251} &  \cca{0.426} &  \cca{0.169} &  \cca{0.242} &  \cca{0.967} &  \cca{0.991} &  \cca{0.979}        \\
ne-en         &  \cca{0.694} &  \cca{0.434} &  \cca{0.760} &  \cca{0.918} &  \cca{0.832} &  \cca{0.746} &  \cca{0.454} &  \cca{0.564} &  \cca{0.192} &  \cca{0.444} &  \cca{0.098} &  \cca{0.161} &  \cca{0.955} &  \cca{0.994} &  \cca{0.974}        \\ \bottomrule
\end{tabular}
}
\caption{Detailed evaluation metric breakdown of all submitted ensemble system on \texttt{test20} test set.\label{tab:breakdown}}
\end{table*}

We find the QE task performance to be quite sensitive to the label balancing factor $\mu$, but there is also no universally optimal value for all language pairs.
Table \ref{tab:lb-factor} shows this behavior for all to-English language pairs.
Notice that while for most of the cases $\mu$ simply controls a trade-off between the performance of \texttt{OK} and \texttt{BAD} outputs, there are also cases such as ro-en where a certain choice of $\mu$ hurts the performance of both classes.
This might be due to a certain label class being particularly hard to fit, thus creating more difficulties with learning when the loss function is designed to skew to this label class.

It should be noted that this label balancing factor does not correlate directly with the ratio of the \texttt{OK} vs. \texttt{BAD} labels in the training set.
For example, to obtain the best performance, ne-en requires $\mu = 3.0$ while en-de requires $\mu = 1.0$, while the \texttt{OK} to \texttt{BAD} ratio for ne-en (2.14:1) is much less skewed compare to en-de (10.2:1).

\paragraph{Detailed Error Breakdown}
We found it hard to develop an intuition for the model performance from the MCC metric.
To further understand which label categories our models struggle with the most, we breakdown the target-side metric into a cross product of \{\texttt{MT}, \texttt{GAP}\} tags and \{\texttt{OK}, \texttt{BAD}\} classes and compute precision, recall and F1-score for each category.
The breakdown is shown in Table \ref{tab:breakdown}.
It can be seen that our model is making the most mistakes with the \texttt{GAP BAD} category, while the category with the least mistakes is the \texttt{GAP OK} category.
Also, note that for MT word tags, the models often seem to suffer more from low precision rather than low recall, while for gaps it is the opposite.

Overall, we see that the highest F1 scores we can achieve for detecting bad MT words or gaps are rarely higher than 0.8, which indicates that there should be ample room for improvement.
It would also be interesting to measure the inter-annotator agreement of these word-level quality labels, in order to get a sense of the human performance we should be aiming for.

%% file: source/conclusion.tex
\section{Conclusion}

In this paper, we present our WMT21 word-level QE shared task submission based on Levenshtein Transformer training and a two-step finetuning process.
We also explore various ways to create synthetic data to build more generalizable systems with limited human annotations.
We show that our system outperforms the OpenKiwi+XLM baseline for all language pairs we experimented with.
Our official results on the blind test set also demonstrate the competitiveness of our system.
We hope that our work can inspire other applications of Levenshtein Transformer beyond the widely studied case of non-autoregressive translation.